\documentclass{article}

\usepackage{arxiv}

\usepackage[utf8]{inputenc} % allow utf-8 input
\usepackage[T1]{fontenc}    % use 8-bit T1 fonts
\usepackage{hyperref}       % hyperlinks
\usepackage{url}            % simple URL typesetting
\usepackage{booktabs}       % professional-quality tables
\usepackage{amsfonts}       % blackboard math symbols
\usepackage{nicefrac}       % compact symbols for 1/2, etc.
\usepackage{microtype}      % microtypography
\usepackage{lipsum}
\usepackage{graphicx}
\usepackage[shortlabels]{enumitem}
\usepackage[numbers]{natbib}
\title{Recognizing more emotions with less data using self-supervised transfer learning}

\author{
  Jonathan Boigne \\
  Wonder Technology\\
  Beijing, China \\
  \texttt{jonathan@wondertech.com.cn} \\
   \And
  Biman Najika Liyanage \\
  Wonder Technology\\
  Beijing, China \\
  \texttt{biman@wondertech.com.cn} \\

  \And
  Ted Östrem \\
  Wonder Technology\\
  Beijing, China \\
  \texttt{ted@wondertech.com.cn} \\

}

\begin{document}

\maketitle

%%%%%%%%%%%%%%%%%%%%%%%%%%%%%%%%%%%%%%%%%%%%%%%%%%%%%%%%%%%%%%%%%%%%%%
\begin{abstract}

We propose a novel transfer learning method for speech emotion recognition allowing us to obtain promising results when only few training data is available. With as low as 125 examples per emotion class, we were able to reach a higher accuracy than a strong baseline trained on 8 times more data. Our method leverages knowledge contained in pre-trained speech representations extracted from models trained on a more general self-supervised task which doesn't require human annotations, such as the wav2vec model. We provide detailed insights on the benefits of our approach by varying the training data size, which can help labeling teams to work more efficiently. We compare performance with other popular methods on the IEMOCAP dataset, a well-benchmarked dataset among the Speech Emotion Recognition (SER) research community.

Furthermore, we demonstrate that results can be greatly improved by combining acoustic and linguistic knowledge from transfer learning. We align acoustic pre-trained representations with semantic representations from the BERT model through an attention-based recurrent neural network. Performance improves significantly when combining both modalities and scales with the amount of data. When trained on the full IEMOCAP dataset, we reach a new state-of-the-art of 73.9{$\%$}  unweighted accuracy (UA).

\end{abstract}

%%%%%%%%%%%%%%%%%%%%%%%%%%%%%%%%%%%%%%%%%%%%%%%%%%%%%%%%%%%%%%%%%%%%%%
\section{Introduction}

Emotion recognition has been gaining traction for the last decade with the interest of providing conversational agents with high EQ when interacting with users. Such systems have applications in healthcare, non-invasive mental health diagnostics and screening, automotive, as well as education. Recognizing emotions has been a challenging task in the speech domain, mainly due to the lack of large-enough labeled datasets to successfully apply proven deep learning techniques, like it has been done in tasks with more resources such as Automatic Speech Recognition (ASR) \cite{hannun_deep_2014, collobert_wav2letter_2016}. This is even more problematic for non-English languages with fewer resources. It is thus important to find alternative ways to train accurate classifiers in situations where data is scarce. Moreover, emotion recognition is a data problem where having a finer emotion classification system leads to datasets of uneven sample sizes or data with multiple labels.

There have been various lines of work attempting to train accurate emotion classifiers with a low amount of labeled data. Some works have proposed to impose stronger restrictions on convolutional layers to better fit raw speech data and prevent overfitting on small data \cite{ravanelli_speaker_2018}. Multi-task learning has also been proposed as a way to mitigate overfitting to a small dataset by simultaneously learning different tasks \cite{xia_multi-task_2017, zhang_attention-augmented_2019}. However, these approaches still require a large enough amount of data to be able to learn efficient filters from scratch or to train all adjacent tasks.

Transfer learning is a growing area of research in deep learning and has the potential to help alleviate this problem of label scarcity. Particularly re-using pre-trained models or representations trained on more general tasks has been successfully applied in other domains than speech \cite{erhan_why_2010}. In computer vision, it is now standard practice to train a deep convolutional model first on the large ImageNet classification dataset and then re-use those weights for fine-tuning on a different vision task where less labeled data is available \cite{huh_what_2016}. In natural language processing, the pre-training of large language models has been widely adopted and led to greatly superior results on many tasks \cite{radford_improving_2018, devlin_bert_2019}.

Despite the clear benefits of pre-training in computer vision and natural language processing, this approach has not yet taken off in the speech domain, mainly because it is still unclear which task is the most suitable for pre-training. Yet recent works have started to show promising results, notably the wav2vec model \cite{schneider_wav2vec_2019} which helped reach a new state-of-the-art for speech recognition by training rich representations from unlabeled data through a contrastive-type of loss. More specifically, for the task of emotion recognition, \cite{lu_speech_2020} obtained strong results by re-using representations from a pre-trained ASR model. However, this approach still requires a large amount of labeled data to first train a good speech transcription model, which makes it impracticable for languages or domains where data is scarce.

In this work, we focus on unsupervised pre-training, which does not require labeled data for learning pre-trained representations, and as such we decide to re-use the wav2vec representations to apply for emotion recognition. We also experiment with combining two different kinds of pre-trained representations for both speech and text. The addition of BERT representations allows us to improve our accuracy by almost 10\%.

%%%%%%%%%%%%%%%%%%%%%%%%%%%%%%%%%%%%%%%%%%%%%%%%%%%%%%%%%%%%%%%%%%%%%%
\section{Approach}

%%%%%%%%%%%%%%%%%%%%%%%%%%%%%%%%%%%%%%%%%%%%%%%%%%%%%%%%%%%%%%%%%%%%%%
\subsection{Speech Emotion Recognition}

The current standard approach to classify emotions from speech is to first extract low-level features, often called low-level descriptors (LLDs) from short frames of speech of duration ranging from 20 to 50 milliseconds, denoted \(\mathbf{x}_{1:T}=(x_1,x_2,...,x_T)\), where \(x_i \in R^n\), \(n\) being the number of features. A high-level aggregation transformation is then applied to convert these frame-level features to an utterance-level representation \(x_{utterance} \in R^n\). Finally, a softmax layer is applied on top of this new global representation to classify the utterance along the possible emotion classes.

Different methods have been used to transform the variable-length sequence of low-level descriptors to a fixed-length representation at the utterance level. Traditionally, statistical aggregation functions were applied to each of the LLDs over the duration of the utterance such as mean, max, variance, etc. These were concatenated into a long feature-vector to represent a single utterance. With the emergence of neural networks as the preferred approach, there have been experiments with various strategies such as recurrent neural networks or attention-based models \cite{mirsamadi_automatic_2017}.

%%%%%%%%%%%%%%%%%%%%%%%%%%%%%%%%%%%%%%%%%%%%%%%%%%%%%%%%%%%%%%%%%%%%%%
\subsubsection{Acoustic features}

Traditionally most works use hand-crafted features that have been proven to work well for other speech-related tasks. Commonly used features include Mel-frequency cepstrum coefficients (MFCCs), zero-crossing rate, energy, pitch, voicing probability, etc. Another line of work has been focused on using raw spectrogram magnitudes, log-Mel spectrograms, or even raw waveform \cite{trigeorgis_adieu_2016, sarma_emotion_2018}. Yet although this works well for tasks where data is sufficient like Automatic Speech Recognition, these methods are still difficult to apply for emotion recognition due to a lack of labeled data.

We suggest that working with pre-trained features trained on raw waveform or raw spectrograms could help alleviate this lack of data, and we suggest using wav2vec features that were trained on large data in a self-supervised fashion.

%%%%%%%%%%%%%%%%%%%%%%%%%%%%%%%%%%%%%%%%%%%%%%%%%%%%%%%%%%%%%%%%%%%%%%
\subsection{Pre-trained Wav2vec Representations}

The wav2vec \cite{schneider_wav2vec_2019} model is made of two simple convolutional neural networks: the encoder network and the context network. The encoder network \(f : X \mapsto Z\) takes raw audio samples \(x_i \in X\) as input and outputs low frequency feature representations \((z_1, z_2, \dots, z_T)\) which encode about 30ms of 16kHz audio every 10ms. The context network \(g : Z \mapsto C\) transforms these low-frequency representations into a higher-level contextual representation \(c_i = g(z_i \dots z_{i-v})\) for a receptive field {v}. The total receptive field after passing through both networks is 210ms for the base version and 810ms for the large version.

The model was trained on about 1,000 hours of unlabeled English speech with a noise contrastive binary classification task, where the objective was to distinguish true future samples from distractors. The motivation behind wav2vec was to learn effective pre-trained representations to be used for Automatic Speech Recognition (ASR). Training a model on the resulting representations allowed the team to outperform Deep Speech 2 \cite{amodei_deep_2016}, the best reported character-based system at the time while using two orders of magnitude less data.

Our assumption is that these representations not only contain relevant information for speech recognition but also para-acoustic information that could be useful for detecting emotions in speech.

%%%%%%%%%%%%%%%%%%%%%%%%%%%%%%%%%%%%%%%%%%%%%%%%%%%%%%%%%%%%%%%%%%%%%%
\subsection{Model Architectures}

\begin{figure}
  \centering
  \includegraphics[width=\textwidth]{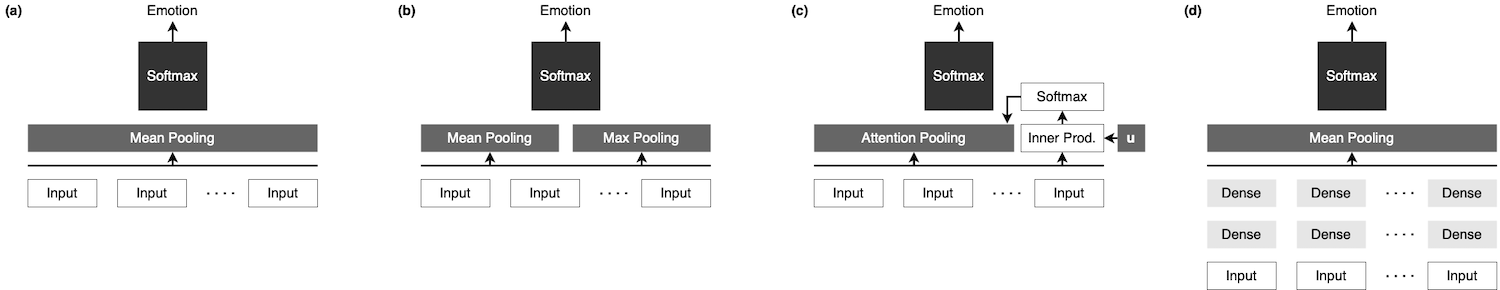}
  \caption{Architectures used for comparing emotion recognition performance using hand-engineered features or pre-trained audio representations. (a) Mean pooling in time. (b) Mean+Max pooling. (c) Attention-weighted pooling. (d) MLP with mean pooling.}
  \label{fig:fig1}
\end{figure}

We compare our approach with a baseline of commonly-used hand-engineered features to see how much more emotional information is contained in the pre-trained representations. To better understand the differences between both kinds of features, we experiment with various model architectures described below and in Figure~\ref{fig:fig1}. Due to our experiments on a very small amount of data, we selected simpler models rather than more complex recurrent neural networks to avoid overfitting. However, we do compare our solution to more advanced baselines trained on more data in Section~\ref{sec:sec4}.

\begin{enumerate}[(a)]
\item
\textbf{Mean pooling:} this approach simply averages each feature along the time dimension. We make the simplifying assumption that the emotional information is constant-enough along time to meaningfully represent the utterance-level overall emotion. We use short-enough audio segments to make sure this is the case.
\item
\textbf{Mean-max pooling:} similar to mean pooling except that we leverage both the mean and the max of features across the time dimension. We then concatenate both vectors before feeding it to the softmax layer. This approach effectively doubles the number of parameters of our model from just mean pooling.
\item
\textbf{Attention pooling:} the model learns by itself a weighted average whose weights are determined by a simple attention mechanism based on logistic regression. This allows the model to learn the most efficient pooling function for the task, including the mean pooling approach above.
\item
\textbf{MLP with pooling:} this last approach allows the model to learn more complex frame-level features before the pooling operation. We build a Multi-Layer Perceptron (MLP) on top of individual frame-level features \(x_1, x_2, \dots, x_T\) before mean pooling across the time dimension. The MLP parameters are shared across time. We scale the number of layers and units to match the model capacity for the two types of features in order to compensate for the smaller dimension of LLD feature vectors compared to wav2vec representations.

\end{enumerate}

%%%%%%%%%%%%%%%%%%%%%%%%%%%%%%%%%%%%%%%%%%%%%%%%%%%%%%%%%%%%%%%%%%%%%%
\subsection{Bimodal Emotion Recognition}
Several works have explored combining information from linguistic features and acoustic features. \citet{yoon_multimodal_2018, heusser_bimodal_2019} combined utterance-level information from audio and textual embeddings before classifying through the last softmax layer. \citet{lu_speech_2020} also used a similar approach to ours except that they used pre-trained representations from an ASR model, which thus already contains semantic information.

In this work, we decide to use the same model as in \cite{xu_learning_2019}, where we align both audio and textual pre-trained representations through an attention mechanism on top of a bidirectional recurrent neural network. The only difference is the replacement of hand-engineered features by wav2vec embeddings and of textual GloVe embeddings \cite{pennington_glove_2014} by BERT embeddings. BERT \cite{devlin_bert_2019} is a Transformer \cite{vaswani_attention_2017} encoder-only model which has been shown to capture meaningful information for text classification tasks. Specifically, it is able to encode word embeddings which take the context into account.

\section{Experimental Setup}

To compare emotion recognition performance between traditional features and embeddings from self-supervised pre-trained models, we use the IEMOCAP dataset \cite{busso_iemocap_2008}, which contains ~12 hours of audiovisual data, including video, speech, motion capture of face, text transcriptions. We use scripted and improvised dialogs from all 5 sessions. We only use the audio (and transcriptions for the bi-modal experiment) for our evaluation.

Similarly to previous works on IEMOCAP, we only use 4 emotion classes (neutral, happy, sad, and angry) where at least 2 annotators agree. We also merge excitement together with happiness, resulting in a total of 5,531 utterances for the full dataset. We evaluate our model using speaker-independent 5-fold cross-validation: for each split, 4 sessions are used for training, and the remaining one is used for testing. We report our model's accuracy as the average of the accuracy on the validation set over all folds. This is consistent with previous works.

We use the pyAudioAnalysis library \cite{giannakopoulos_pyaudioanalysis_2015} to extract a set of 34 commonly-used features as in \cite{xu_learning_2019}. For wav2vec representations, we use the large version of the model since they have a wider receptive field which is more likely to contain emotion-salient features. For both audio representations, we crop all utterances to a maximum duration of 5 seconds. For BERT embeddings, we use the base model and extract embeddings from the provided IEMOCAP transcription files with the HuggingFace library \cite{wolf_huggingfaces_2019}. We take embeddings from the second-to-last layer since these contain more general information and are less tied to BERT's particular training objective.

We train our models with a small batch size of 16 and we use the Adam optimizer with learning rate \(10^{-4}\). We don't use any form of early stopping. We add dropout regularization for all models after each layer.

%%%%%%%%%%%%%%%%%%%%%%%%%%%%%%%%%%%%%%%%%%%%%%%%%%%%%%%%%%%%%%%%%%%%%%
\section{Results}
\label{sec:sec4}
%%%%%%%%%%%%%%%%%%%%%%%%%%%%%%%%%%%%%%%%%%%%%%%%%%%%%%%%%%%%%%%%%%%%%%
\subsection{Training on Few Data}

We compare generalization performance in a setting where training data is very scarce. We use only 500 examples for training, 125 per class. By using only a simple mean pooling and a softmax layer, we were able to reach 56.7\% unweighted accuracy (UA) thanks to the wav2vec representations. This is 5.2\% more than the same model trained on standard hand-engineered features. Table~\ref{tab:table_500} report our results across different models and compare them with hand-engineered features. Our best model was able to reach an unweighted accuracy of 58.5\%, higher than the same model trained on 8 times more data using hand-engineered features.

Interestingly, our results almost match a more advanced Bi-RNN model \cite{mirsamadi_automatic_2017}, trained on the entire dataset, that allows each frame to attend to the entire utterance (see Table~\ref{tab:table_final}). In comparison, wav2vec representations only have a receptive field of about 810ms, less than 1/6 of the maximum utterance length of 5 seconds.

\begin{table}[t]
\caption{Unweighted accuracy when training on only 500 examples using different features and model architectures}
\begin{center}
\begin{tabular}{lccc}
\toprule
Model & LLD Features & Pre-trained Representations &  Improvement \\
\midrule
Mean pooling & 49.7{$\%$}  & 54.9{$\%$} & \textbf{+5.2\%} \\
Mean+Max pooling & 49.9{$\%$}  & 57.4{$\%$} & \textbf{+7.5\%} \\
Attention pooling & 49.7{$\%$}  & 55.3{$\%$} & \textbf{+5.6\%} \\
MLP with pooling & 51.6{$\%$}  & 58.5{$\%$} & \textbf{+6.9\%} \\
\bottomrule
\end{tabular}
\label{tab:table_500}
\end{center}
\end{table}

\subsection{Performance Scaling}

\begin{figure}
  \centering
  \includegraphics[width=\textwidth]{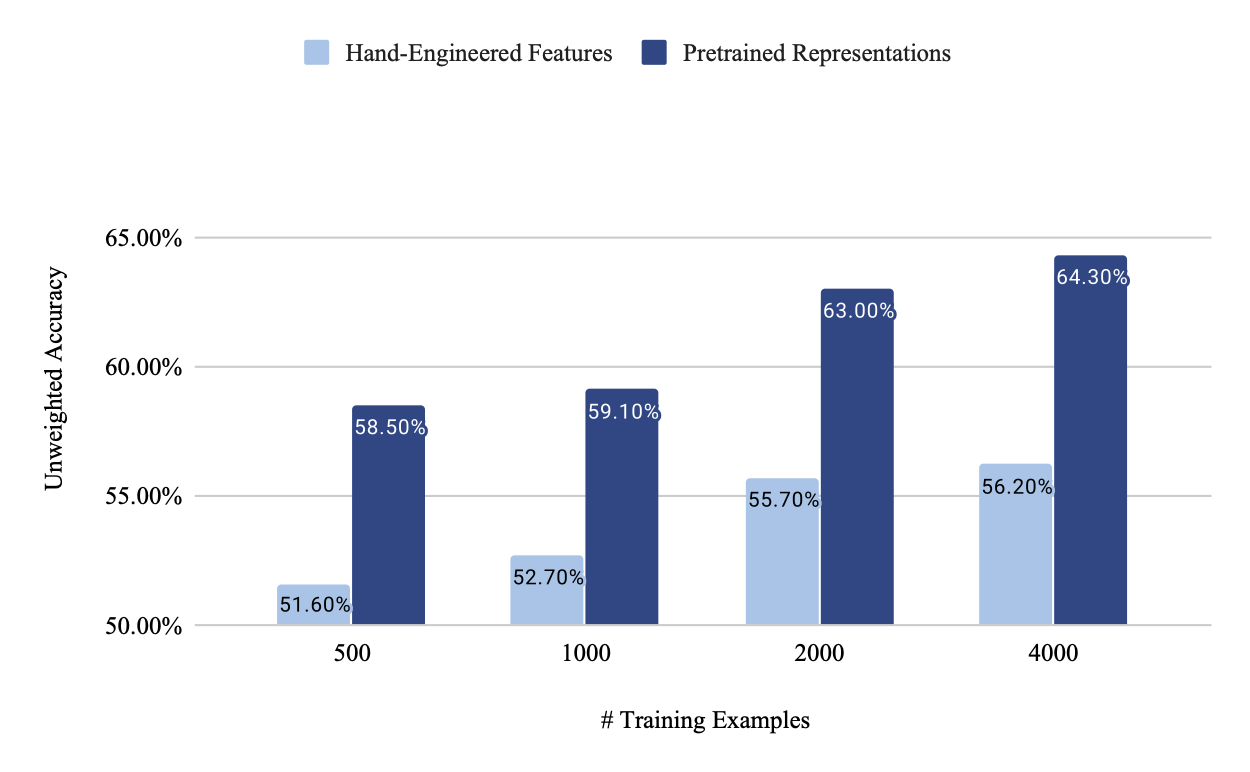}
  \caption{Evolution of performances for our best model (MLP with mean pooling) in function of the amount of training data}
  \label{fig:fig2}
\end{figure}

In a second phase, we progressively increase the amount of training data on our best model (MLP with pooling). Figure~\ref{fig:fig2} shows the evolution of the unweighted accuracy performance as we increase the amount of data. The UA increases in a log fashion with respect to the number of training examples.

Starting from 2,000 examples, our approach strongly outperforms other acoustic-only baselines that we benchmark against (see Table~\ref{tab:table_final}). The models we compare with were not only trained on more than twice the data, but they also present much more complex architectures compared to our simple MLP with pooling.

\subsection{Bi-Modal Transfer Learning}

Last, we experiment with combining pre-trained embeddings for both audio and text. We align wav2vec representations and sub-words embeddings from BERT through an attention-based recurrent neural network to align both representations in time, similar to \cite{xu_learning_2019}. The resulting model is much larger than previous ones, and to avoid over-fitting we only train it on the full dataset. We report the unweighted accuracy in Table~\ref{tab:table_final}, where our approach outperforms other models and reach a new state-of-the-art unweighted accuracy of 73.9\%.

%%%%%%%%%%%%%%% begin table   %%%%%%%%%%%%%%%%%%%%%%%%%%
\begin{table}[t]
\caption{Performance of our acoustic-only and bi-modal models when scaled on the full IEMOCAP dataset. We report results from previous state-of-the-art and often-cited works for comparison.}
\begin{center}

\begin{tabular}{lll}
& & \\ % put some space after the caption
\toprule
Model & Features & UA \\
\midrule
& \multicolumn{2}{l}{Acoustic only} \\
\cmidrule(l){2-3}
Bi-LSTM with attention pooling \cite{mirsamadi_automatic_2017} & Low-level descriptors & 58.8{$\%$} \\
CNN with Bi-LSTM \cite{satt_efficient_2017} & Raw spectrograms & 59.4{$\%$} \\
TDNN with LSTM and attention \cite{sarma_emotion_2018} & Raw waveform & 60.7{$\%$} \\
\textbf{MLP with mean pooling} & \textbf{Pre-trained wav2vec representations} & \textbf{64.3{$\%$}} \\
\midrule
& \multicolumn{2}{l}{Acoustic and Textual} \\
\cmidrule(l){2-3}
Bi-LSTM with attention alignment \cite{xu_learning_2019} & LLDs and GloVe embeddings & 70.9{$\%$} \\
Bi-LSTM with multi-head self-attention \cite{lu_speech_2020} & pre-trained ASR representations & 72.6{$\%$} \\
\textbf{Bi-LSTM with attention alignment}  & \textbf{Pre-trained wav2vec and BERT representations} & \textbf{73.9{$\%$}} \\

\bottomrule
\end{tabular}
\label{tab:table_final}
\end{center}
\end{table}

\section{Conclusion}

In this paper, we compare performance for emotion recognition when using pre-trained embeddings learned in a self-supervised setting. We demonstrate the superior performance and sample-efficiency of our technique compared to identical models with commonly-used hand-engineered features. Our model is able to reach a higher accuracy with 8 times less data than if it was trained from scratch in a supervised setting.

We report performance as we scale up the training data, and build a final model on two modalities: audio and text. Both modalities use pre-trained features from self-supervised models, and we reach a new state-of-the-art unweighted accuracy of 73.9\%.

\bibliography{references}

\end{document}